%% file: paperX.tex
\documentclass{article}
\usepackage{spconf}
\usepackage{multirow}

\usepackage{graphicx} 
\usepackage{amsmath}
\usepackage{bbm}
\usepackage{ifthen}
\usepackage{adjustbox}
\usepackage{xkeyval}
\usepackage{tikz}
\usepackage{calc}
\usepackage{amssymb}
\usepackage{multirow}
\usepackage{paralist}
\usepackage{comment}
\usepackage{booktabs}       
\usepackage[hidelinks]{hyperref}
\usepackage{multirow}
\usepackage[normalem]{ulem}
\useunder{\uline}{\ul}{}
\usepackage{xspace}
\usepackage{colortbl}
\usepackage{lipsum}

\newcommand\blfootnote[1]{%
  \begingroup
  \renewcommand\thefootnote{}\footnote{#1}%
  \addtocounter{footnote}{-1}%
  \endgroup
}

\usepackage[capitalize]{cleveref}
\crefname{section}{Sec.}{Secs.}
\Crefname{section}{Section}{Sections}
\Crefname{table}{Table}{Tables}
\crefname{table}{Tab.}{Tabs.}

\def\etal{\emph{et al. }}
\def\eg{\emph{e.g. }}
\def\ie{\emph{i.e. }}

\newcommand{\convabb}{SRE\xspace}
\newcommand{\convname}{SRE-Conv\xspace}

\title{SRE-Conv: Symmetric Rotation Equivariant Convolution for Biomedical Image Classification}
\name{
Yuexi~Du$^{1}$, 
Jiazhen~Zhang$^{1}$, 
Tal~Zeevi$^{1}$, 
Nicha~C.~Dvornek$^{1,2}$,
John~A.~Onofrey$^{1,2,3}$
}
\address{
    Departments of 
    $^1$Biomedical Engineering, 
    $^2$Radiology \& Biomedical Imaging, \\
    $^3$Urology, 
    Yale University, New Haven, CT, USA \\
}

\begin{document}

\maketitle
\begin{abstract}
Convolutional neural networks (CNNs) are essential tools for computer vision tasks, but they lack traditionally desired properties of extracted features that could further improve model performance, e.g., rotational equivariance. 
Such properties are ubiquitous in biomedical images, which often lack explicit orientation. 
While current work largely relies on data augmentation or explicit modules to capture orientation information, this comes at the expense of increased training costs or ineffective approximations of the desired equivariance. 
To overcome these challenges, we propose a novel and efficient implementation of the Symmetric Rotation-Equivariant (\convabb) Convolution (\convname) kernel, designed to learn rotation-invariant features while simultaneously compressing the model size. The \convname kernel can easily be incorporated into any CNN backbone. We validate the ability of a deep \convabb-CNN to capture equivariance to rotation using the public MedMNISTv2 dataset (16 total tasks). 
\convname-CNN demonstrated improved rotated image classification performance accuracy on all 16 test datasets in both 2D and 3D images, all while increasing efficiency with fewer parameters and reduced memory footprint.
The code is available on \url{https://github.com/XYPB/SRE-Conv}.
\end{abstract}
\blfootnote{\textcopyright~2025 IEEE. Personal use of this material is permitted. Permission from IEEE must be obtained for all other uses, in any current or future media, including reprinting/republishing this material for advertising or promotional purposes, creating new collective works, for resale or redistribution to servers or lists, or reuse of any copyrighted component of this work in other works.}

\input{tex/introduction.tex}

\input{tex/method.tex}

\input{tex/results.tex}

\input{tex/discussion.tex}

\input{tex/acknowledgements.tex}

\bibliographystyle{IEEEbib}
\bibliography{references}

\end{document}

%% file: tex/introduction.tex
\section{Introduction}
\label{sec:introduction}

Convolution layers play a pivotal role in the successful application of deep learning methods to computer vision and biomedical image analysis tasks, where the learned convolutional kernel weights demonstrate frequency- and orientation-specific responses to images~\cite{krizhevsky2012imagenet}. 
The significance of convolution kernels lies in capturing local patterns while preserving spatial relationships within images.
Translational equivariance is a key advantage of convolution, enabling networks to recognize features regardless of their position within an image, and their shared weights further optimize computation by reducing the number of parameters and enhance model efficiency.
However, convolutional kernels exhibit limitations with respect to rotational equivariance, which makes convolutional neural networks (CNNs) sensitive to rotations, hindering their effectiveness in scenarios where rotational variations are prevalent, \eg histopathology imaging of cells. 
A common practice to address this problem is to use geometric data augmentation approaches, such as rotation and reflection, during model training; however, these methods strictly increase computational training costs~\cite{quiroga2020revisiting} and can result in performance degradation due to interpolation artifacts.
Furthermore, global geometric augmentation of the data may not cover all possible transformations in the application, especially if the rotation occurs locally, \eg a cell rotates locally while the overall orientation of the tissue is unchanged in a histopathology image. 
Equivariant convolution kernels allow the extraction of features equivariant to rotations, reflections, and translations, ensuring the model's capability to generalize across different orientations.

A variety of approaches have been proposed to achieve equivariance in CNNs.
\textit{Orientation-aware neural networks} learn orientation information actively during training from the data and use the learned information to re-align the images to their standardized orientation~\cite{jaderberg2015spatial} or learn this information by aligning all image gradients to a similar orientation~\cite{hao2022gradient}.
\textit{Rotation-encoded neural networks} encode pre-defined rotation transformations using circular harmonics~\cite{worrall2017harmonic}, steerable filters~\cite{weiler2019general}, group-equivalent operations~\cite{cohen2016group}, or actively rotate the filters during convolution ~\cite{zhou2017oriented}.
\textit{Rotation-equivariant coordinate systems} ensure rotational equivariance by transforming the input data to a different coordinate system, \eg cyclic coordinate systems~\cite{mo2022ric} or polar coordinates~\cite{paletta2022spin}.
\textit{Weight symmetric convolution} methods, to which our approach belongs, explicitly encode the convolution kernel weights to have symmetric properties such as rotational equivariance~\cite{dudar2019use,fuhl2021rotated}, but past performance in these cases was limited due to small kernel sizes, \eg 3$\times$3, that hindered the model's ability to learn expressive features. Moreover, most of the previous works focus on 2D images while ignoring 3D image applications.

In response to these challenges, we present a straightforward yet effective design for the convolutional kernel that achieves translation, rotation, and reflection equivariance. 
Inspired by classical orientation-independent edge detection in image analysis using the Laplacian of Gaussian~\cite{Marr1980-pn}, our proposed symmetric rotation-equivariant (\convabb) convolutional (\convname) kernel leverages the properties of symmetric kernels and the translation-equivariant nature of convolution operations~\cite{dudar2019use,fuhl2021rotated}. 
The central symmetric kernel ensures not only rotation and reflection equivariance but also reduces the number of trainable parameters compared to conventional convolutional layers as well as simultaneously enabling expanded receptive field sizes~\cite{Luo2017-sy}.
Importantly, \convname maintains the same computational complexity as traditional convolutions without introducing additional computation overhead and can be seamlessly integrated into existing deep learning architectures and frameworks.
The novel contributions of this work include:
\begin{inparaenum}[(i)]
    \item a formulation to create symmetric kernels of arbitrary size and controllable symmetric pattern;
    \item incorporating large \convabb kernels into a variety of deep neural network architectures (both 2D and 3D);
    \item validating \convname on a diverse set of 2D and 3D biomedical image classification tasks available in the public MedMNISTv2~\cite{yang2023medmnist}
    datasets (16 tasks in total); and 
    \item making our code available for public use.
\end{inparaenum}
Our experimental results show that \convname significantly enhances the performance and robustness of rotation-equivariant biomedical image classification, outperforming baseline methods in both accuracy and parameter efficiency.

%% file: tex/method.tex
\section{Methods}
\label{sec:methods}

\input{fig/fig_method}

\subsection{Symmetric Rotation-Equivariant (\convabb) Convolution}

We employ a centrally symmetric kernel~\cite{Marr1980-pn} for equivariance (\cref{fig:method}).
The \convname kernel is parameterized using non-overlapping discrete circular bands based on their Euclidean distance from the kernel center, where each band represents one trainable parameter. 
This arrangement ensures local equivariance within the kernel operation space.
However, global equivariance is ensured by the translation-equivariant property of convolution operations.
By sliding the \convname kernel across the entire image, the locally equivariant kernel convolves with each local image patch. 
Consequently, the resulting feature map maintains rotational equivariance, even after global rotation and translation of the input, as all values remain consistent with respect to a rotation transformation. 

We define a $k \times k$ 2D \convname kernel with $C$ channels (we denote $C$ for $C_\text{in}$ and $C_\text{out}$ for simplicity) as a matrix $K\in\mathbb{R}^{[C, k, k]}$ (\cref{fig:method}).
We parameterize $K$ using $b = \lfloor k/2\rfloor + 2$ discrete non-overlapping symmetric bands.
We use a sparse binary index matrix $M_I\in\mathbb{R}^{[b,k,k]}$ to map the $b$ trainable weight parameters $\Theta\in\mathbb{R}^{[C,b]}$ to the kernel space, \ie $K = M_I \Theta$.
We reshape the index matrix $M_I$ to flatten its spatial dimension to be $M^f_I\in\mathbb{R}^{[b,k^2]}$ that stacks the individual bands to obtain the final binary index matrix.
We note that $M^f_I$ is the same for all $C$ channels and is fixed during the training.
To create ``circular'' shaped symmetric kernels, we zero out the columns in $M^f_I$ that correspond to the four corners.
To create the \convname's bands, we first calculate a Euclidean distance matrix $D\in\mathbb{R}^{[k, k]}$ with respect to the center of the kernel and then split it into $b$ equal bands.
We build the binary index matrix $M^f_I$ by assigning 1 to each row-column index that has equal distance values in $D$.
We compute $M^f_I$ once during kernel initialization. 
This parameterization allows the gradient to back-propagate to the $\Theta$ parameter matrix when updating the model. 
The bias term remains the same since it does not influence the rotational equivariance.
We can extend 2D \convname to 3D without loss of generality.

\subsection{\convname Parameter Efficiency}
Compared to a standard $k\times k$ convolution kernel, the \convabb kernel reduces the dimensionality of the trainable parameters from $O(C\cdot k^2)$ to $O(C\cdot b) = O(C\cdot k/2 + 2) = O(C\cdot k)$. 
This approach enables efficient parameterization of large kernels with large effective receptive fields~\cite{Luo2017-sy}.
For kernels with the same size, \convname uses the same number of computational operations as standard convolution.
At inference, we pre-compute the full convolutional kernel $K = M_I \Theta$ using the trained weights to eliminate the extra computation for matrix multiplication. 
Therefore, our \convabb kernel will have the same number of floating operations (FLOPs) as traditional convolution during inference.
Parameter efficiency improves further when we extend to 3D \convname.

\input{fig/tab_2d}

\subsection{\convabb-CNN Architecture}

We construct a fully convolutional CNN (FCN) using our \convname layers (\convabb-CNN).
The FCN architecture, which does not include any reshaping or flattening operations on the spatial dimensions, allows the network to maintain rotational equivariance in the rotated feature maps across layers.
Global adaptive pooling after the final feature extraction layer ensures that the classifier head operates over rotation-equivariant features.
Due to the central symmetric property of \convabb, we have local rotational equivariance around the center of a convolved patch. 
However, if the input is rotated and a convolutional stride greater than 1 is used, it may cause the kernel to convolve at a different position, which results in non-equivariant feature maps. 
Therefore, in contrast to existing CNNs that use convolutions with a stride greater than 1 to downsample the image, we use an equivariant pooling layer followed by a 1$\times$1 convolutional layer with stride 1.

%% file: fig/fig_method.tex
\begin{figure}[t]
    \centering
    \includegraphics[width=1.0\columnwidth]{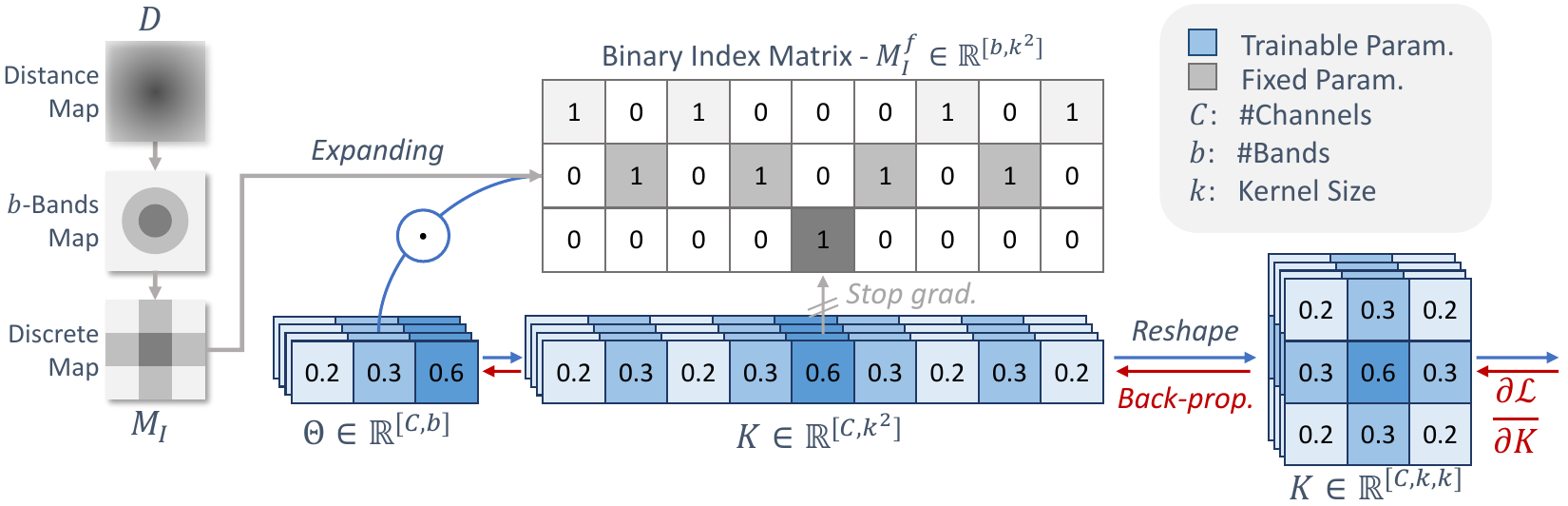}
    \vspace{-1.8\baselineskip}
    \caption{{\bf Proposed \convname Kernel.} 
    To construct an \convname kernel of arbitrary size $k\times k$ with $b$ discrete symmetric bands and $C$ channels, we multiply a small trainable weight matrix $\Theta$ with a pre-defined binary index matrix $M^f_I$ (fixed during training). An \convname layer can be used in place of any standard convolution layer.}
    \label{fig:method}
    \vspace{-1.0\baselineskip}
\end{figure}

%% file: fig/tab_2d.tex
\begin{table*}[t]
\centering
\caption{\textbf{2D MedMNISTv2 Evaluation.} Classification accuracy of each model on the original test set (Orig.), rotated test set (Rot.), and reflected test set (Ref.) for 10 2D MedMNISTv2~\cite{yang2023medmnist} datasets. R18 refers to the ResNet-18 architecture and W16 refers to WideResNet-16. Size indicates the model's size in terms of the number of parameters. Mem. indicates the GPU memory cost for each model. 
We highlight the top result in bold and the second-best with an underline. 
}
\label{tab:2d}
\resizebox{0.95\linewidth}{!}
{
\footnotesize
\begin{tabular}{lcccc|ccc|ccc|ccc|ccc}
\toprule
\multicolumn{1}{c}{\multirow{2}{*}{\textbf{Method}}} & \multicolumn{1}{c}{\multirow{2}{*}{~~\textbf{Size}~~}} & \multicolumn{3}{c|}{\textbf{Pathology}} & \multicolumn{3}{c|}{\textbf{Chest}} & \multicolumn{3}{c|}{\textbf{Derma}} & \multicolumn{3}{c|}{\textbf{OCT}} & \multicolumn{3}{c}{\textbf{Pneumonia}} \\ 
\cmidrule(l){3-17} 
\multicolumn{1}{c}{} & \multicolumn{1}{c}{} & ~Orig.~ & ~Rot.~ & ~Ref.~ & ~Orig.~ & ~Rot.~ & ~Ref.~ & ~Orig.~ & ~Rot.~& ~Ref.~ & ~Orig.~ & ~Rot.~ & ~Ref.~ & ~Orig.~ & ~Rot.~ & ~Ref.~ \\ \midrule
R18~\cite{he2016deep} & 11.2M & 89.5 & {\ul 80.2} & 89.4 & 94.5 & 94.3 & 94.5 & \textbf{78.8} & 64.5 & 74.7 & {\ul 76.4} & 33.5 & 51.6 & \textbf{93.3} & {\ul 77.4} & 73.1 \\ 
ORN-R18~\cite{zhou2017oriented} & 11.5M & 86.9 & 73.5 & 87.2 & 94.7 & 94.8 & {\ul 94.7} & 74.9 & 57.2 & 73.3 & 70.6 & 31.7 & 60.7 & 85.4 & 67.0 & 77.8 \\
G-R18~\cite{cohen2016group} & 11.6M & {\ul 92.0} &  78.7 & {\ul 91.8} & 94.0 & 91.7 & 93.5 & 71.5 & 61.9 & 71.4 & 73.7 & 30.4 & 62.0 & 84.0 & 69.9 & 77.2  \\
H-R18~\cite{worrall2017harmonic} & 17.3M & 82.1 & 60.8 & 82.1 & 94.8 & 94.7 & 94.5 & 74.1 & 67.9 & 72.7 & 72.5 & 33.5 & 63.4 & 86.5 & 72.3 & 77.5 \\
E(2)-W16~\cite{weiler2019general} & {\ul 10.8M} & \textbf{92.2} & 73.4 & \textbf{92.0} & {\ul 94.8} & {\ul 94.8} & {\ul 94.7} & {\ul 77.7} & {\ul 74.3} & \textbf{77.4} & 75.3 & {\ul 48.0} & {\ul 75.2} & 88.0 & 74.2 & {\ul 88.0} \\
RIC-R18~\cite{mo2022ric} & 11.2M & 89.5 & 71.7 & 89.5 & 94.1 & 94.0 & 93.8 & 74.5 & 68.7 & 73.3 & 71.8 & 42.6 & 64.9 & 85.5 & 70.1 & 75.4 \\  \midrule
\rowcolor[HTML]{EFEFEF}  %
\convabb-R18 & \textbf{3.9M} & 82.2 & \textbf{82.9} & 82.2 & \textbf{94.8} & \textbf{94.8} & \textbf{94.8} &  75.7 & \textbf{75.3} & {\ul 75.7} & \textbf{76.8} & \textbf{52.2} & \textbf{76.8} & {\ul 90.6} & \textbf{90.2} & \textbf{90.6} \\ \midrule \midrule
\multicolumn{1}{c}{\multirow{2}{*}{\textbf{Method}}} & \multicolumn{1}{c}{\multirow{2}{*}{~~\textbf{Mem.}~~}} & \multicolumn{3}{c|}{\textbf{Retina}} & \multicolumn{3}{c|}{\textbf{Breast}} & \multicolumn{3}{c|}{\textbf{Blood}} & \multicolumn{3}{c|}{\textbf{Tissue}} & \multicolumn{3}{c}{\textbf{OrganA}} \\ \cmidrule(l){3-17} 
\multicolumn{1}{c}{} & \multicolumn{1}{c}{} & ~Orig.~ & ~Rot.~ & ~Ref.~ & ~Orig.~ & ~Rot.~ & ~Ref.~ & ~Orig.~ & ~Rot.~& ~Ref.~ & ~Orig.~ & ~Rot.~ & ~Ref.~ & ~Orig.~ & ~Rot.~ & ~Ref.~ \\ \midrule
R18~\cite{he2016deep} & 3.8GB & {\ul 50.8} & 48.3 & 48.8 & \textbf{ 87.8} & 74.5 & 75.3 & 95.8 & {\ul 92.9} & 95.8 & 69.0 & 43.4 & 63.9 &  93.3 & 46.6 & 53.1 \\ 
ORN-R18~\cite{zhou2017oriented} & 8.8GB & 46.8 & 40.4 & 45.5 & 86.5 & 64.0 & 75.6 & 95.3 & 85.5 & 94.9 & 61.4 & 31.5 & 59.7 & 91.8 & 33.3 & 62.0 \\
G-R18~\cite{cohen2016group} & 13.8GB & 46.3 & 47.7 & {\ul 49.3} & 85.2 & {74.8} & {\ul 80.6} & 96.6 & 88.3 & 96.3 & 67.2 & 46.3 &  65.8 & \textbf{95.8} & 46.1 & {\ul 70.5} \\
H-R18~\cite{worrall2017harmonic} & 4.5GB & 47.3 & 48.3 & 49.1 & 76.9 & 63.7 & 71.4 & 96.1 & 92.5 & 95.5 & 67.0 & 45.5 & 65.0 & 91.4 & 42.4 & 63.4 \\
E(2)-W16~\cite{weiler2019general} & 18.5GB & 46.0 & 44.6 & 47.3 & {\ul 87.2} & 66.4 & \textbf{80.8} & \textbf{97.7} & 84.5 & \textbf{97.7} & \textbf{72.3} & {\ul 48.7} & {\ul 70.5} & {\ul 94.5} & {\ul 56.9} & 73.3 \\
RIC-R18~\cite{mo2022ric} & {\ul 3.6GB} & 49.3 & {\ul 48.9} & 48.1 & 84.0 & {\ul 75.0} & 75.0 & {\ul 96.7} & 91.3 & {\ul 96.4} & 67.1 & 34.4 & 64.8 & 92.7 & 42.9 & 64.6\\ \midrule  \rowcolor[HTML]{EFEFEF}%
\convabb-R18 & \textbf{2.9GB} & \textbf{52.3} & \textbf{51.6} & \textbf{52.3} & 79.5 & \textbf{80.39} & 79.5 & 96.0 & \textbf{95.9} & 96.0 & {\ul 71.2} & \textbf{55.6} & \textbf{71.2} & 76.9 & \textbf{71.5} & \textbf{76.9} \\  
\bottomrule
\end{tabular}
}
\vspace{-1.0\baselineskip}
\end{table*}

\input{fig/tab_3d}

%% file: fig/tab_3d.tex
\begin{table*}[t]
\centering
\caption{\textbf{3D MedMNISTv2 Evaluation.} Classification accuracy of each model on both the original test set (Orig.) and rotated test set (Rot.) from 6 3D MedMNISTv2~\cite{yang2023medmnist} datasets. 
We highlight the top result in bold. }
\label{tab:3d}
\resizebox{0.95\linewidth}{!}
{
\begin{tabular}{lcccc|cc|cc|cc|cc|cc}
\toprule
\multicolumn{1}{c}{\multirow{2}{*}{\textbf{Method}}} & \multicolumn{1}{c}{\multirow{2}{*}{~~\textbf{Size}~~}} & \multicolumn{1}{c}{\multirow{2}{*}{~~\textbf{Mem.}~~}} & \multicolumn{2}{c|}{\textbf{Organ}} & \multicolumn{2}{c|}{\textbf{Nodule}} & \multicolumn{2}{c|}{\textbf{Adrenal}} & \multicolumn{2}{c|}{\textbf{Fracture}} & \multicolumn{2}{c|}{\textbf{Vessel}} & \multicolumn{2}{c}{\textbf{Synapse}} \\ \cmidrule(l){4-15} 
\multicolumn{1}{c}{} & \multicolumn{1}{c}{} & \multicolumn{1}{c}{} & ~Orig.~ & ~Rot.~ & ~Orig.~ & ~Rot.~ & ~Orig.~ & ~Rot.~ & ~Orig.~ & ~Rot.~ & ~Orig.~ & ~Rot.~ & ~Orig.~ & ~Rot.~ \\ \midrule
R3D-18~\cite{tran2018closer} & 33.2M & 10.4GB & \textbf{95.7} & 56.1 & 85.5 & 84.6 & 76.9 & 74.1 & 54.6 & 43.5 & \textbf{97.6} & 91.8 & 75.6 & 71.9 \\
R2plus1D-18~\cite{tran2018closer} & 31.3M & \textbf{6.1GB} & 93.8 & 52.2 & 85.5 & 84.1 & 77.9 & 71.2 & 49.2 & 43.6 & 96.6 & 91.8 & 78.1 & 76.6 \\ \midrule \rowcolor[HTML]{EFEFEF}
\convabb-R3D-18 & \textbf{2.6M} & 7.7GB & 74.8 & \textbf{65.2} & \textbf{88.7} & \textbf{88.3} & \textbf{84.2} & \textbf{80.4} & \textbf{58.8} & \textbf{45.7} & 92.9 & \textbf{93.2} & \textbf{84.1} & \textbf{82.8} \\ \bottomrule
\end{tabular}
}
\vspace{-1.0\baselineskip}
\end{table*}

%% file: tex/results.tex
\section{Experiments and Results}
\label{sec:results}

\subsection{Experimental Setup}

\paragraph*{Datasets:}

To evaluate the effectiveness of our model across various medical imaging modalities, we validate \convabb-CNN on 
the public MedMNISTv2~\cite{yang2023medmnist} classification dataset.
MedMNISTv2 comprises 12 2D and 6 3D medical imaging datasets, covering a wide spectrum of clinical applications.
Benchmarking on such a diverse and intricate dataset helps to validate the model's ability to generalize to various tasks. 
We skip the OrganC and OrganS datasets because they are repeated with OrganA and Organ3D.

\paragraph*{Baselines:}

For 2D evaluation, we use the conventional \textbf{ResNet18 (R18)}~\cite{he2016deep} as the baseline for CNNs that are not rotation equivariant. We compare our \textbf{\convabb-R18} with 5 state-of-the-art rotation equivariant baselines with the ResNet-style backbone: \textbf{ORN-R18}~\cite{zhou2017oriented}, \textbf{Group-R18 (G-R18)}~\cite{cohen2016group}, \textbf{Harmonic-R18 (H-R18)}~\cite{worrall2017harmonic},
 \textbf{E(2)-W16}~\cite{weiler2019general} where W16 is WideResNet-16, and \textbf{RIC-R18}~\cite{mo2022ric}. 
For 3D evaluation, we choose \textbf{R3D-18}~\cite{tran2018closer} and \textbf{R2plus1D-18}~\cite{tran2018closer} as our baselines. 
Our \textbf{\convabb-R3D-18} shares the same architecture as the R3D-18~\cite{tran2018closer} but using our \convname3D layers.  
Official implementations are used for evaluation.

\paragraph*{Evaluation Metrics:}

We evaluate model performance for each dataset by computing classification accuracy on:
\begin{inparaenum}[(1)]
    \item the \textbf{original test} set without rotation;  
    \item the \textbf{rotated test} set (rotate by 10{\textdegree} increments for 2D and by 30{\textdegree} increments about each axis for 3D); and
    \item the \textbf{reflected test} set (horizontal and vertical flips in 2D).
\end{inparaenum}
We assess significant differences ($\alpha$=$0.05$) between models by computing two-tailed paired t-tests comparing across the independent datasets.

\paragraph*{\convabb-CNN Implementation Details:}

As our model naturally has fewer parameters than the conventional convolutional layer, we choose to use a larger kernel size in our model, \ie $[9,9,5,5]$ for each stage in the ResNet architecture for both 2D and 3D evaluation. 
We use the same training and testing settings for all MedMNISTv2~\cite{yang2023medmnist} datasets. We train the models with the SGD optimizer and cosine annealing scheduler with an initial learning rate of $2\times10^{-2}$. 
The model is trained for 100 epochs with a batch size of 128 for 2D images and 4 for 3D images.
We use the cross-entropy loss for all multi-class classification tasks, and we use the binary cross-entropy loss for the multi-label classification ChestMNIST dataset. All models are implemented with PyTorch and trained with an NVIDIA A5000 GPU.
As is standard practice for studies on equivariant feature learning~\cite{zhou2017oriented,cohen2016group,worrall2017harmonic,weiler2019general,mo2022ric}, no geometric data augmentation is applied during training in order to 
demonstrate the full capabilities of equivariant learning without introducing confounding effects.

\input{fig/fig_vis}
\subsection{Results}

\paragraph*{2D MedMNISTv2 Evaluation:} 
\convabb-R18 outperforms all baselines on the rotated test set with a notable gap (\cref{tab:2d}), especially for datasets where images have no specific orientation, \ie Pathology, Derma, Retina, and Blood. Our model also performs well even in the original test set without rotation, demonstrating either best or second best accuracy in 5 of 10 tasks. 
It achieves an impressive performance using only $\sim30\%$ of the model size of the corresponding standard R18 CNN and the minimum GPU memory usage (\cref{tab:2d}). We note that our model failed on the OrganA dataset, where the images are strictly oriented in the same anatomical direction. The symmetric constraint makes it hard to adapt to the orientation-specific prior in the dataset, which results in a lower performance on the original test set. Still, overall on the 2D datasets, our method does not perform statistically differently from the conventional ResNet18 on the original test sets ($p$=0.37) even though ResNet18 is $\sim3\times$ larger in model size. 
Notably, our method performs significantly better on the rotated test sets compared with every baseline model ($p$$<$0.005). Furthermore, all baseline models showed significantly degraded performance on the rotated test set compared to the original test set ($p$$<$0.05), while only our method demonstrates no significant difference ($p$=0.11) in performance between the original and rotated test sets. A similar improvement was found in the reflect evaluation where our method ensures strict reflection equivariance with no performance loss.

\paragraph*{3D MedMNISTv2 Evaluation:} 
Similar to the 2D
results, we observe the same success in the rotated evaluation of our model for all 6 datasets (\cref{tab:3d}), with significantly improved accuracy compared with both baselines ($p$=0.02 for each model). 
\convabb-R3D-18 underperforms on the original test set of datasets with strong orientation prior, \eg Organ.
However, overall our model does not perform significantly differently on the original datasets compared with the baselines ($p$=0.93 for each model). Yet, it is notable that our model achieves this level of performance with \emph{less than 10\%} of the parameters (2.6M) compared with the baselines (33.2M and 32.2M, respectively). This highlights the parameter efficiency advantage of our method in 3D data, where the number of parameters always increases linearly w.r.t. kernel size.

\paragraph*{Qualitative Results:} We visualize feature map of the standard R18 and \convabb-R18 on rotated test images selected from two 2D MedMNISTv2 datasets (\cref{fig:featuremap}). 
We unrotate each feature map from the rotated input back to the original input orientation.
A circular mask is applied for visualization purposes only.
Our model produces consistent feature maps across all rotations, while the feature map of the conventional CNN changes dramatically even with a small rotation. This illustrates why our model can perform better on medical imaging datasets with no specific orientation.

%% file: fig/fig_vis.tex
\begin{figure}[!t]
    \centering
    \includegraphics[width=1.0\columnwidth]{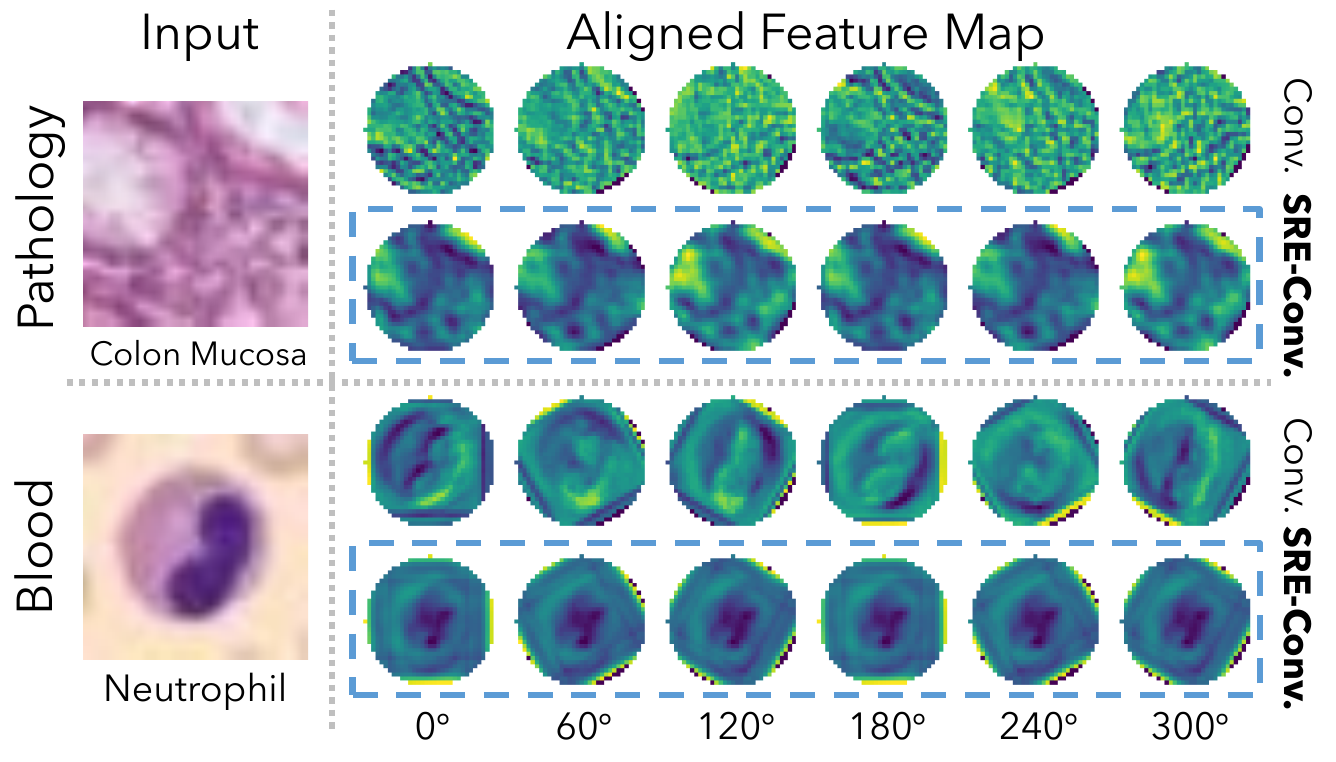}
    \vspace{-2.0\baselineskip}
    \caption{{\bf Feature Map Visualization.} 
    We visualize feature maps of the conventional R18 (Conv.) and our \convname R18 models on two 2D MedMNISTv2 datasets. We rotate input images in $60^\circ$ increments, extract the averaged first layer feature map, and unrotate the feature map back to align with the original input. We apply a circle mask for visualization.} 
    \label{fig:featuremap}
    \vspace{-1.0\baselineskip}
\end{figure}

%% file: tex/discussion.tex
\section{Discussion and Conclusion}
\label{sec:discussion}

By incorporating rotational equivariance directly into the convolutional kernel design, deep learning models can learn equivariant features that are invariant to rotation and reflection, enhancing their robustness to real-world scenarios where objects may vary in orientation. 
Our results demonstrate that our \convname layers with equivariant kernels improve accuracy and reliability across a wide range of biomedical imaging applications.
\convabb-CNNs demonstrated the best classification accuracy for all 16 clinical tasks on rotated test sets in the MedMNISTv2
datasets and even demonstrated the best or second-best accuracy in 9 of 16 tasks on the original test set 
compared to other state-of-the-art rotation equivariant baseline models.
Furthermore, our \convname representations are parameter efficient as they scale linearly with kernel size compared to exponentially for traditional kernels, and our experiments demonstrate that \convabb-CNNs can maintain performance on various datasets despite fewer parameters. 
The equivariant features learned by \convname (Fig.~\ref{fig:featuremap}) maintain their identity under transformations, are crucial for robust and reproducible feature learning, and have tremendous potential to provide robust and reproducible imaging biomarkers.
While the performance of our approach can be limited in some datasets with natural orientation, \eg OrganMNIST, we aim to address this by designing a model inspired by visual cortex processing that combines \convname and conventional convolutions with orientation encoding.

%% file: tex/acknowledgements.tex
\paragraph*{Acknowledgements}
No funding was received to conduct this study. The authors have no relevant financial or non-financial interests to disclose.

\paragraph*{Compliance with Ethical Standards}

This research study was conducted retrospectively using human subject data made available in open access by Yang~\etal~\cite{yang2023medmnist}. Ethical approval was not required as confirmed by the license attached with the open-access data.